\documentclass[letterpaper, 10 pt, conference]{ieeeconf}  % Comment this line out if you need a4paper

\IEEEoverridecommandlockouts                              % This command is only needed if 
\usepackage{graphicx} % for pdf, bitmapped graphics files
\usepackage{amsmath} % assumes amsmath package installed
\usepackage{amsfonts}
\usepackage{bbold}

\title{\LARGE \bf
Belief-Aided Navigation using Bayesian Reinforcement Learning for Avoiding Humans in Blind Spots
}

\author{Jinyeob Kim$^{1}$, Daewon Kwak$^{1}$, Hyunwoo Rim$^{2}$, and Donghan Kim$^{2}$*% <-this % stops a space
\thanks{$^{1}$Author with Department of Artificial Intelligence, College of Software, Kyung Hee University, Yongin, Republic of Korea
        {\tt\small wls2074@khu.ac.kr, kdw1181@khu.ac.kr}}%
\thanks{$^{2}$Author with Department of Electronic Engineering (AgeTech-Service Convergence Major), College of Electronics \& Information, Kyung Hee University, Yongin, Republic of Korea
        {\tt\small hwlim2000@khu.ac.kr}}%
\thanks{*Corresponding author
        {\tt\small donghani@khu.ac.kr}}% <-this % stops a space
}

\begin{document}

\maketitle
\thispagestyle{empty}
\pagestyle{empty}

%%%%%%%%%%%%%%%%%%%%%%%%%%%%%%%%%%%%%%%%%%%%%%%%%%%%%%%%%%%%%%%%%%%%%%%%%%%%%%%%
\begin{abstract}

Recent research on mobile robot navigation has focused on socially aware navigation in crowded environments. However, existing methods do not adequately account for human–robot interactions and demand accurate location information from omnidirectional sensors, rendering them unsuitable for practical applications. In response to this need, this study introduces a novel algorithm, BNBRL+, predicated on the partially observable Markov decision process framework to assess risks in unobservable areas and formulate movement strategies under uncertainty. BNBRL+ consolidates belief algorithms with Bayesian neural networks to probabilistically infer beliefs based on the positional data of humans. It further integrates the dynamics between the robot, humans, and inferred beliefs to determine the navigation paths and embeds social norms within the reward function, thereby facilitating socially aware navigation. Through experiments in various risk-laden scenarios, this study validates the effectiveness of BNBRL+ in navigating crowded environments with blind spots. The model's ability to navigate effectively in spaces with limited visibility and avoid obstacles dynamically can significantly improve the safety and reliability of autonomous vehicles.

\end{abstract}

%%%%%%%%%%%%%%%%%%%%%%%%%%%%%%%%%%%%%%%%%%%%%%%%%%%%%%%%%%%%%%%%%%%%%%%%%%%%%%%%
\section{INTRODUCTION}

Recent investigations have explored socially aware navigation using mobile robots within densely populated settings. Traditional reactive-based methodologies such as optimal reciprocal collision avoidance (ORCA) [1, 2] and social force (SF) [3] find differentiating between humans and inanimate objects, as well as comprehending social norms [4], difficult, often resulting in aggressive path planning strategies. To surmount these obstacles, deep reinforcement learning (DRL) [5] has been advanced for executing complex tasks [6-8].

Nonetheless, numerous algorithms assume a Markov decision process (MDP) [9] by assuming the robot’s capability to discern human positions omnidirectionally, which may not be feasible because of the physical limitations on the sensors in sim2real scenarios, thereby curtailing their real-world applicability.

To address this limitation, the present study introduces a pioneering approach for ensuring safe navigation within a partially observable Markov decision process (POMDP) framework. This method accounts for the potential hazards in areas not covered by sensors to ensure secure navigation. It leverages a single robot with a constrained field of view (FoV) within the POMDP context. 

Notably, a state-of-the-art (SOTA) algorithm [10] for POMDP navigation employs a variational autoencoder (VAE) to infer the positions of unseen individuals through analysis of human–human (HH) interactions. However, its utility diminishes considerably in environments without human–human interactions [10], and its robustness is further challenged in real-world contexts owing to difficulties in discerning human–human interactions amid environmental noise. 

Contrarily, this study advances a methodology that forecasts future trajectories solely based on the observed locations of the individuals [11], independent of human–human interactions. By utilizing belief algorithms to estimate the positions of non-visible individuals predicated on this forecasted data, this approach not only surpasses previous research in robustness but also demonstrates superior performance in environments where all sensors are powered off, as evidenced by the experiments detailed in Section V.

Another state-of-the-art (SOTA) algorithm has been investigated for its application in multirobot systems within a decentralized partially observed semi-Markov decision process (dec-POSMDP) environment [6, 12]. This approach [6] facilitated socially aware navigation by employing multiple robots with a limited FoV in a dec-POSMDP setting, thereby enabling the detection of both humans and robots.

However, the use of multiple robots introduces complexities due to the necessity of mutual detection, coupled with the challenge of recognizing individuals beyond the FoV of the robots. Similarly, a SOTA algorithm employing a single robot encounters obstacles in achieving socially aware navigation in densely populated areas, attributed to its limitations in processing information outside the FoV of the robot [8, 13].

Conversely, this study pioneers in facilitating uncertainty-aware navigation through the deployment of Bayesian neural networks (BNNs) predicated on belief information to ascertain the probability of human collision risks in zones beyond sensor coverage [14]. Moreover, it effectively addresses the freezing robot dilemma by identifying key features of belief–belief (BB) and robot–belief (RB) interactions via a spatio temporal interaction graph (sti-graph) [15-17]. Additionally, incorporation of social norms into reward mechanisms via DRL enables socially aware navigation with superior performance.

Consequently, the contributions of this research are manifold. First, an algorithm has been devised and executed, which leverages trajectory prediction and belief algorithms within a POMDP framework, relying exclusively on the location data of individuals. This method proves more pragmatic and resilient compared to those based in MDP environments by considering areas beyond the reach of sensor detection, thereby mitigating collision risks and improving stability. Second, by harnessing belief information, a novel network termed as the Belief-aided Bayesian Reinforcement Learning + attention module (BNBRL+) has been introduced. This innovation integrates the network utilized in prior studies [1] with BNNs, empowering a single robot with a restricted FoV to deduce the potential interactions among the robot, humans, and beliefs via a sti-graph. BNBRL+ further integrates social norms into the reward structure, facilitating socially aware navigation.

\section{RELATED WORKS}

\subsection{Inferencing Uncertainty Using Beliefs}

Navigating within a POMDP framework presents multifaceted sources of uncertainty, particularly when traversing areas lacking sensor data, which can cause the “freezing robot” phenomenon [16]. To mitigate this, belief-based methodologies have been deployed.

A study [18] addressed the challenges of intersection scenarios by integrating DRL with ensemble techniques for the computation of beliefs. Multiple networks were trained across diverse datasets to generate a variety of beliefs regarding the same input for navigation purposes. However, this strategy is based on the assumption that most hazards (e.g., vehicles and pedestrians) exhibit predictable patterns (e.g., pedestrians crossing roads via crosswalks).

Further explorations [19] derived beliefs by examining agent interactions in contexts where hazards are not directly observable, such as occluded obstacles on roads. Another research [10] identified occluded agents in densely populated environments through agent interactions, which is an approach designated as social perception. Despite these advancements, these algorithms encounter robustness challenges, attributed to inaccuracies such as false positives and negatives. Moreover, their performance may degrade in environments lacking interactive elements or sensor data. 

\subsection{Bayesian Neural Networks}

As a belief represents a synthesis of uncertain information, it requires a probabilistic framework to effectively manage this ambiguity. 

Conventional DNNs are prone to overfitting and exhibit overconfidence in their predictions [14]. To mitigate these issues, BNNs have been introduced using the Bayes by Backprop algorithm [14]. This technique imbues weight parameters with uncertainty by treating them as probability distributions, thereby enhancing the model robustness and improving generalization by acknowledging variability in training data. Consequently, BNNs outperform DNNs in generalization capability and ensure more reliable model behavior by incorporating uncertainty. 

In the field of BNN applications, a study [20] underscored the advantages of Bayesian reinforcement learning (BRL) in achieving high performance with limited data samples while navigating uncertainties in real-world settings. This research advocates for the use of Bayesian approaches and priors to deduce posterior distributions in uncharted territories, including mixed observability Markov decision processes [21]. Thus, it efficiently inferred uncertainty information related to gestures, achieving optimal performance. 

In this study, the challenge of navigating crowded environments with inherent risks such as limited FoV and intermittent visibility is addressed. Predicting the movements of individuals in densely populated areas, where movements deviate from regular patterns poses a significant challenge, surpassing those encountered in previous studies [11, 14]. To this end, we employed DRL and model-free methodologies, supplemented by a trajectory predictor to estimate future positions of individuals. 

In addition, we utilized a trajectory predictor to forecast the future positions of individuals and probabilistically inferred the posteriors using belief and BNNs [11, 14]. This approach does not rely on social perception and presents a new method suitable for the POMDP framework. This methodology is further detailed in Section IV.

\section{BACKGROUND}

\subsection{Partially Observable Markov Decision Process}

POMDP is an environment in which the agent does not have complete knowledge of all the states. It is defined by the following elements: continuous state $S$, initial state $S_0$, action spaces $A$, transition probability $P$, observation space $\mathrm{\Omega}$, observation $O$, reward $R$, discount factor $\gamma$. In each episode, the agent starts from the initial state $S_0$ and at each time $t$, selects $a_t \in A$ based on the observation $o_t\in\mathrm{\Omega}$ detected by the sensors, according to the policy: $\pi(a_t|o_t)$ = $\mathbb{P}[A=a_t|S=o_t]$. The policy $\pi$ is updated with the objective of maximizing the discounted cumulative reward: $\mathbb{E}\left[\sum_{t=0}^{\infty}{\gamma^tR(s_t,a_t,s_{t+1})}\right],$ where $\gamma\in[0,1]$ denotes a discount factor. Additionally, the environment transitions to the subsequent state $s_{t+1}\in S$ according to $P(s_{t+1}|s_t)$.

\subsection{Spatio temporal Interaction Graph}

The sti-graph aids robots to comprehend human behavior and think broadly like humans [8, 15]. This module comprises two stages: a spatio interaction graph and a temporal interaction graph. In addition, the sti-graph includes nodes $\mathcal{V}_t$ representing agents and edges $\mathcal{E}_t$ representing features between the nodes at time step $t : \mathcal{G}_t=(\mathcal{V}_t,\mathcal{E}_t)$. The sti-graph analyzes $\mathcal{E}_t$ at the same time step $t$, extracting the latent features between $\mathcal{V}_t$ within the range of the sensor, thereby capturing the overall movements of the agents. However, the temporal interaction graph integrates and analyzes $\mathcal{G}_t$ of adjacent time steps $t$, enabling the robot to yield a broad perspective similar to that of humans.

\section{METHODS}

   \begin{figure*}[t]%
      \centering
      \includegraphics[width=1.0\textwidth]{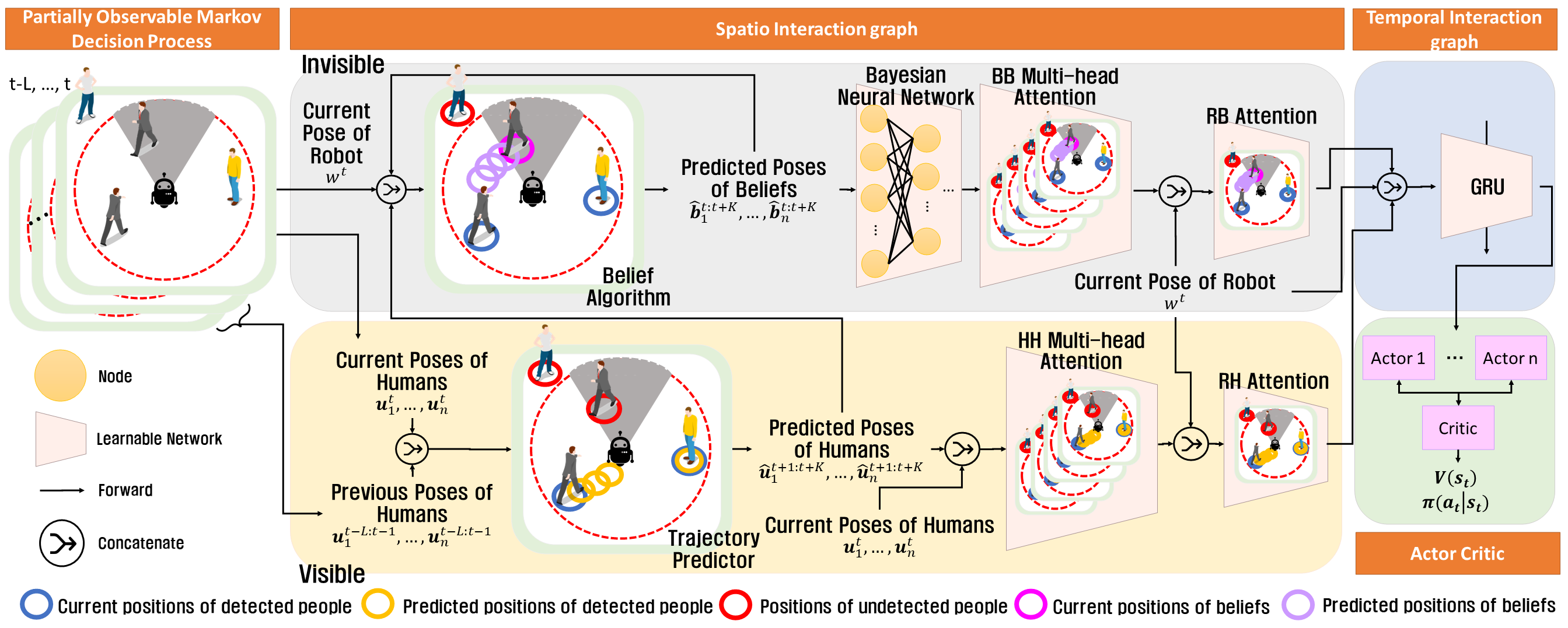}
      \caption{The structure of BNBRL+}
      \label{figurelabel}
   \end{figure*}

The structure of the proposed network is described in this section. This network uses and enhances the architecture proposed in [8].

It consists of four stages (orange boxes in Fig. 1). The first stage (POMDP) involves the robot recognizing surrounding individuals through sensors. The second stage (spatio interaction graph) is divided into parts processing visible human data and handling invisible data. The third stage (temporal interaction graph) employs a gated recurrent unit (GRU) [17], and the fourth stage utilizes an actor-critic [23].

\subsection{Partially Observable Markov Decision Process}

In this research, we considered an environment exclusively populated by robot and humans, with the robot's FoV restricted to 270°. This limitation precluded the robot from acquiring information about agents or any out-of-range data within the depicted gray area in Fig. 1. Consequently, the positional information of the $i-th$ individual at time step $t$, denoted as $\mathit\mathbf{u}_i^t=(p_x^i,p_y^i)$, the relative positional information $U_t=[\mathit\mathbf{u}_1^t,\ \mathit\mathbf{u}_2^t,\ldots,\mathit\mathbf{u}_n^t]$ up to the $n-th$ detected person can be obtained through sensors. In addition, the past positional information of the $i-th$ person from time steps $t-L$ to $t-1$ is defined as $\mathit\mathbf{u}_i^{t-L:t-1}$. Furthermore, information regarding the robot is defined as $\mathit\mathbf{w}^t=[\mathit\mathbf{p}_{rob}^t,\mathit\mathbf{v}_{rob}^t,\mathit\mathbf{g}^t,v_{max},\theta,\rho]$, including the robot's positional information $\mathit\mathbf{p}_{rob}^t=(p_x,p_y)$, velocity information $\mathit\mathbf{v}_{rob}^t=(v_x,v_y)$, destination location $\mathit\mathbf{g}^t=(g_x,g_y)$, maximum velocity $v_{max}$, heading angle $\theta$, and the radius length $\rho$. Consequently, following the POMDP framework, the observation at time step $t$ comprises $o_t=[\mathit\mathbf{w}^t,\mathit\mathbf{u}_1^t,\ \mathit\mathbf{u}_2^t,\ldots,\mathit\mathbf{u}_n^t]$.

\subsection{Spatio-interaction Graph (Visible)}

We utilized the Gumbel social transformer (GST) to effectively predict the future locations of individuals by extracting latent features among agents [11]. The GST considers the positional information of the detected individual as input from time step $t-L$ to $t-1$, denoted as $U_{t-L:t-1}$ = [$\mathit\mathbf{u}_1^{t-L:t-1},\ \mathit\mathbf{u}_2^{t-L:t-1},\ldots,\mathit\mathbf{u}_n^{t-L:t-1}$]. Additionally, it receives mask data $m_i^t$, indicating whether the $i-th$ individual is detected at time step $t$, and defines the mask for all individuals as $M_t=[m_1^t,m_2^t,\ldots,m_n^t]$. Furthermore, it uses $M_{t-L:t-1}$ as the input. The output predicts the future locations from time step $t+1$ to $t+K$.

\begin{equation}
{\hat{U}}_{t+1:t+K}=GST\left(U_{t-L:t-1},M_{t-L:t-1}\right).
\end{equation}

Therefore, using (1), we obtain the predicted location information ${\hat{U}}_{t+1:t+K}$=[${\hat{\mathit\mathbf{u}}}_1^{t+1:t+K},\ {\hat{\mathit\mathbf{u}}}_2^{t+1:t+K},\ldots,{\hat{\mathit\mathbf{u}}}_n^{t+1:t+K}$] corresponding to the yellow circle in Fig. 1.

Subsequently, the correlation between human–human (HH) and robot–human (RH) interactions is derived using an attention mechanism [24].

The correlation of human–human is learned through self-MHA using multiple independent attentions to learn features from various perspectives. The current and predicted locations of individuals [$U_t,{\hat{U}}_{t+1:t+K}$] are linearly transformed through the learnable weights $W_j^Q,W_j^K,W_j^V$ of the $j-th$ attention, and query $Q$, key $K$, and value $V$ are computed as follows:

\begin{equation}
\begin{gathered}
X=[U_t,{\hat{U}}_{t+1:t+K}], \\
Head_j=Projection_j\left(X\right)=XW_j^Q,XW_j^K,XW_j^V.
\end{gathered}
\end{equation}

Each head bears a different weight, enabling it to learn distinct features or interactions. Thus, the attention score for the $j-th$ head was computed as follows:

\begin{equation}
Attn_j=Softmax\left(\frac{Q_jK_j^T}{\sqrt{d_k}}\right)V_j.
\end{equation}

$Q_j,K_j,V_j$ represent the query, key, and value tensors of the $j-th$ head, respectively; $d_k$ denotes the dimensionality of the key tensor.

Finally, the output of each head is combined to generate the final output:

\begin{equation}
v_{HH}=Concat\left(Attn_1,Attn_2,\ldots,Attn_H\right)W^O.
\end{equation}

where $H$ represents the total number of heads in the attention mechanism, and $W^O$ denotes the weight applied to the final linear transformation of the output tensor from the multihead attention. Furthermore, the result of the human–human interaction at time step $t$ is defined as $\mathit\mathbf{v}_{HH}$, enabling the model to comprehend inputs from various perspectives and discern specific patterns or relationships. Additionally, to expedite and enhance the convergence of the human–human multihead attention, a mask $M_t$ is employed for human detection to consider only detected humans. 

Second, robot–human attention is derived by utilizing cross attention, a mechanism for learning the relationships between two features. While the fundamental structure remains consistent with (2) and (3). $Q$ and $V$ are projected from $\mathit\mathbf{v}_{HH}$, and $K$ is projected from $\mathit\mathbf{w}^t$ to compute the attention scores.

\begin{equation}
Q=\mathit\mathbf{v}_{HH}W^Q,K=\mathit\mathbf{w}^tW^K,V=\mathit\mathbf{v}_{HH}W^V.
\end{equation}

$W^Q,W^K,W^V$ are trainable weights for $Q$, $K$, and $V$, respectively. Subsequently, through the same process as in (3), $\mathit\mathbf{v}_{RH}$ can be obtained; the resulting output represents the robot–human interaction.

\subsection{Spatio-interaction Graph (Invisible)}

In [8], the position was not tracked when there were no agents in the FoV of the robot. However, the belief algorithm continues to predict the position even when humans are not visible. Denoting the belief of the $i-th$ person at time step $t$ by ${\hat{\mathit\mathbf{b}}}_i^t\notin O$, the belief algorithm updates ${\hat{B}}_{t:t+K_{new}}= [{\hat{\mathit\mathbf{b}}}_1^{t:t+K},{\hat{\mathit\mathbf{b}}}_2^{t:t+K} ,\ldots,{\hat{\mathit\mathbf{b}}}_n^{t:t+K}]$ given $\mathit\mathbf{w}_t, {\hat{U}}_{t+1:t+K}, {\hat{B}}_{t:t+K_{old}}$, as depicted in Fig. 1.

   \begin{figure}[b]%
      \centering
      \includegraphics[scale=0.7]{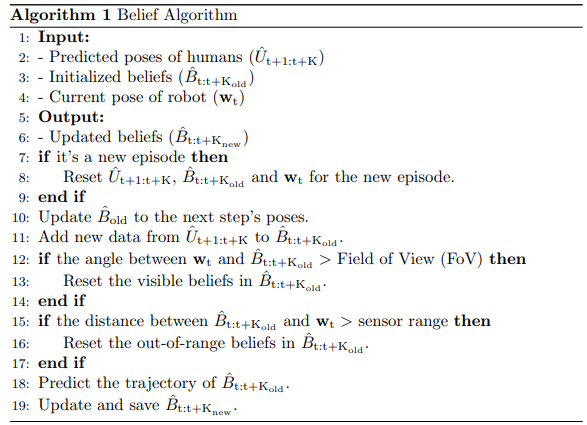}
   \end{figure}

Algorithm 1 presents the pseudocode for the belief algorithm. ${\hat{B}}_{t:t+K}$ is updated based on ${\hat{U}}_{t+1:t+K}$ when ${\hat{U}}_{t+1:t+K}$ no longer tracks the position of the agent owing to the limited FoV (line 11). Subsequently, if ${\hat{B}}_{t:t+K}$ enters the FoV of the robot or falls out of range, the data are reset (lines 12–17). Assuming ${\hat{B}}_{t:t+K}$ traverses along the trajectory according to GST (line 10), it is applied to (1) based on the position at time step $t$ and past belief positions (lines 18–19):

\begin{equation}
{\hat{B}}_{t+1:t+K}=GST\left(B_{t-L:t-1},N_{t-L:t-1}\right).
\end{equation}

where $N_t$ represents the masked data for the beliefs. In addition, the input data are initialized when a new episode starts (lines 7–9).

The belief represents data that do not rely on sensor-detected results and can impose a high dependence on GST. Therefore, an error exists between the actual trajectory and the trajectory of belief: ${\hat{B}}_{t:t+K}\neq{\hat{U}}_{t:t+K}$, implying that belief cannot be applied to DNNs with overconfidence issues [14]. Furthermore, individuals in this experiment alter their direction and destination with a certain probability.

In this study, a framework considering the uncertainty (BNN) was applied [14]. Given training data $\mathcal{D}$, the BNN evaluates the posterior probability $p(w|\mathcal{D})$ for weights $w$ [16]. However, as it can not compute the probability considering all of the error from the unpredictable occurrences, $p(w|\mathcal{D})$ cannot be feasibly obtained. Therefore, it is approximated as similar to the posterior through the history of the episode. Therefore, this was inferred using the Bayesian rule:

\begin{equation}
p\left(w\middle|\mathcal{D}\right)=\frac{p(\hat{Y}|{\hat{B}}_{t:t+K},w)\cdot p(w)}{p(\hat{Y}|{\hat{B}}_{t:t+K})}.
\end{equation}

$\hat{Y}$ denotes the predicted distribution of the unknown state. $p(w)$ serves as the prior, representing the initial distribution of the weights, whereas $p(\hat{Y}|{\hat{B}}_{t:t+K},w)$ denotes the likelihood, indicating how well the data are modeled. $p(\hat{Y}|{\hat{B}}_{t:t+K})$ acts as evidence, normalizes, and provides the probability of the data being independent of the parameters. However, predicting the evidence is typically infeasible, similar to the use of an infinite ensemble of DNNs.

Thus, a two-layer BNN with weight distributions was constructed to infer $p\left(w\middle|\mathcal{D}\right)$ regarding uncertainty in belief and derive the probabilistic features of belief. Subsequently, setting $X={\hat{B}}_{t:t+K}$, the calculation of $\mathit\mathbf{v}_{RB}$ is conducted using (2)–(5).

As a result, this section incorporates uncertain information about unseen space into its path planning process. Moreover, by focusing on high-risk beliefs through the attention mechanism, it navigates in a way that further enhances safety.

\subsection{Temporal Interaction Graph}

Temporal interaction graph was achieved by computing the hidden state of the GRU at time step $t$, using $\mathit\mathbf{v}_{RH}, \mathit\mathbf{v}_{RB}, \mathit\mathbf{w}^t$, and the previous hidden state $h^{t-1}$ at time step $t-1$ as the inputs.

\begin{equation}
h^t=GRU(h^{t-1},{[\mathit\mathbf{v}}_{RH},\mathit\mathbf{w}^t,\mathit\mathbf{v}_{RB}]).
\end{equation}

Incorporating information from adjacent time steps widens the sights of the robot.

\subsection{Actor Critic}

The simulation was executed across 16 parallel environments. The actor updates the policy $\pi(a_t|s_t;\theta_{pol})$, whereas the critic updates the value function $V(s_t;\phi)$. $\theta_{pol}$ represents the parameters of the policy $\pi$, and $\phi$ represents the parameters of the value function. The actor and critic engage in collaborative learning [23], with the proximal policy optimization (PPO) algorithm facilitating the optimization process [25].

\subsection{Reward Function}

We construct a reward function $R(s_t,a_t)$ to train the actions of the robot in the desired direction. $R(s_t,a_t)$ is subdivided into six reward functions: $R_{goal}\left(s_t\right)=10$ for when the robot reaches the destination, $R_{col}\left(s_t\right)=-10$ for collisions, $R_{disc}\left(s_t\right)$ for when the distance between the robot and a person is within a certain range, $R_{pred}\left(s_t\right)$ and $R_{bel}(s_t)$ when the robot is above ${\hat{U}}_{t+1:t+K}$ or ${\hat{B}}_{t+1:t+K}$, and finally $R_{pot}\left(s_t\right)$ based on the variations in the distance to the destination.

$R_{disc}\left(s_t\right)$ represents a reward function designed to maintain a certain distance from a person according to proxemics [4]. In this study, a penalty is incurred when within 0.5m of a person $(s_t\in S_{danger\ zone})$. In addition, a transformable Gaussian reward function (TGRF) [26] based on a Gaussian function incorporating social norms was applied. The reward function is expressed as follows:

\begin{equation}
R_{disc}\left(s_t\right)=TGRF\left(w_{disc},\ 0,\ \sigma_{disc};d_{min}\right).
\end{equation}

$w_{disc}$ and $\sigma_{disc}$ denote the weights and sigma of the TGRF, respectively, set to 0.25 and 0.2. $d_{min}$ denotes the minimum distance between the robot and person.

$R_{pred}\left(s_t\right)$ denotes the penalty incurred when the robot is at ${\hat{U}}_{t+1:t+K}$ representing the estimated future state of the humans.

\begin{equation}
\begin{gathered}
R_{\text{pred}}^i(s_t) = \min_{k=1,\ldots,K} \left(\mathbb{1}_i^{t+k}\frac{R_{\text{col}}}{2^k}\right), \\
R_{\text{pred}}(s_t) = \min_{i=1,\ldots,n} R_{\text{pred}}^i(s_t).
\end{gathered}
\end{equation}

$\mathbb{1}_i^{t+k}$ indicates whether the $i-th$ human represents the presence of a robot at the predicted location at time step $t+k$. Similarly, $R_{bel}\left(s_t\right)$ represents the penalty incurred if the robot is in ${\hat{B}}_{t+1:t+K}$. The formula used is as follows:

\begin{equation}
\begin{gathered}
R_{\text{bel}}^i(s_t) = \min_{k=1,\ldots,K} \left(\mathbb{1}_i^{t+k}\frac{R_{\text{col}}}{2^k}\right)\gamma_{bel}, \\
R_{\text{bel}}(s_t) = \min_{i=1,\ldots,n} R_{\text{bel}}^i(s_t).
\end{gathered}
\end{equation}

$\gamma_{bel}\in[0,1]$ represents the discount factor for belief, with ${\hat{\mathit\mathbf{b}}}_{t+1:t+K}$ being discounted according to the number of tracked steps. In this study, the value was set as 0.9.

$R_{pot}\left(s_t\right)$ denotes the reward function for the distance to the destination, which is expressed by the following formula:

\begin{equation}
    R_{pot}\left(s_t\right)=1.5(-d_{goal}^t+d_{goal}^{t-1}).
\end{equation}

The term $d_{goal}^t$ represents the distance between the robot and destination at time step $t$. Therefore, $R(s_t,a_t)$ is defined as follows:

\begin{equation}
\begin{gathered}
R(s_t,a_t) = \\
\begin{cases}
+10, & \text{if } s_t \in S_{\text{goal}} \\
-10, & \text{if } s_t \in S_{\text{collision}} \\
R_{\text{disc}}(s_t),  & \text{if } s_t \in S_{\text{danger zone}} \\
R_{\text{pred}}(s_t) + R_{\text{bel}}(s_t) + R_{\text{pot}}(s_t), & \text{otherwise}
\end{cases}
\end{gathered}
\end{equation}

Intuitively, the reward function considers a robot's awareness of people's locations, future trajectories, and beliefs, guiding it to select actions that enable it to reach its destination quickly.

\section{EXPERIMENTS}

\subsection{Experimental Environment and Ablation Study}

The experiment is conducted within a 2D environment where the robot adheres to holonomic kinematics, implying it faces no limitations regarding its position and orientation. It is outfitted with a LiDAR system capable of a maximum range of 5m and a FoV of 270°. The radius of the robot is set at $\rho$ = 0.3m, and its maximum velocity is $v_{max}$$ = 1.0m/s$. 

The simulation area measures 12m × 12m, inhabited by 20 individuals who navigate according to the ORCA, engaging in mutual interactions. Notably, the movements of these individuals are unaffected by the position of the robot. These individuals are characterized by random sizes, ranging from 0.3 to 0.5m, and velocities, spanning from 0.5 to 1.5m/s. In addition, humans may change their destination according to a certain probability. 

An episode concludes upon the robot's arrival at its destination, a collision with an individual, or failure to reach the destination within a designated time limit. Diverse outcomes are ensured through the application of random seeds, with the training phase encompassing approximately 20,800 episodes (approximately $1\times{10}^7$ time steps).

The evaluations are performed over 500 distinct episodes, considering metrics such as success rate (SR), navigation time (NT), path length (PL), and intrusion ratio (ITR). 

This research contrasts the newly developed method, BNBRL+, against SOTA learning-based methodologies including GST + HH Attn [8] and DSRNN [13], along with the reactive-based approach ORCA [1, 2], which serve as benchmark comparisons. Two additional models were introduced to demonstrate the effectiveness of BNBRL+.

\begin{itemize}
\item BNDNN: Model replacing BNN with DNN in Fig. 1.
\item BNBRL: Model excluding belief–belief (BB) and robot–belief (RB) attention mechanisms in Fig. 1.
\end{itemize}

\subsection{Result 1 (Limited FoV)}

   \begin{figure*}[t]%
      \centering
      \includegraphics[width=1.0\textwidth]{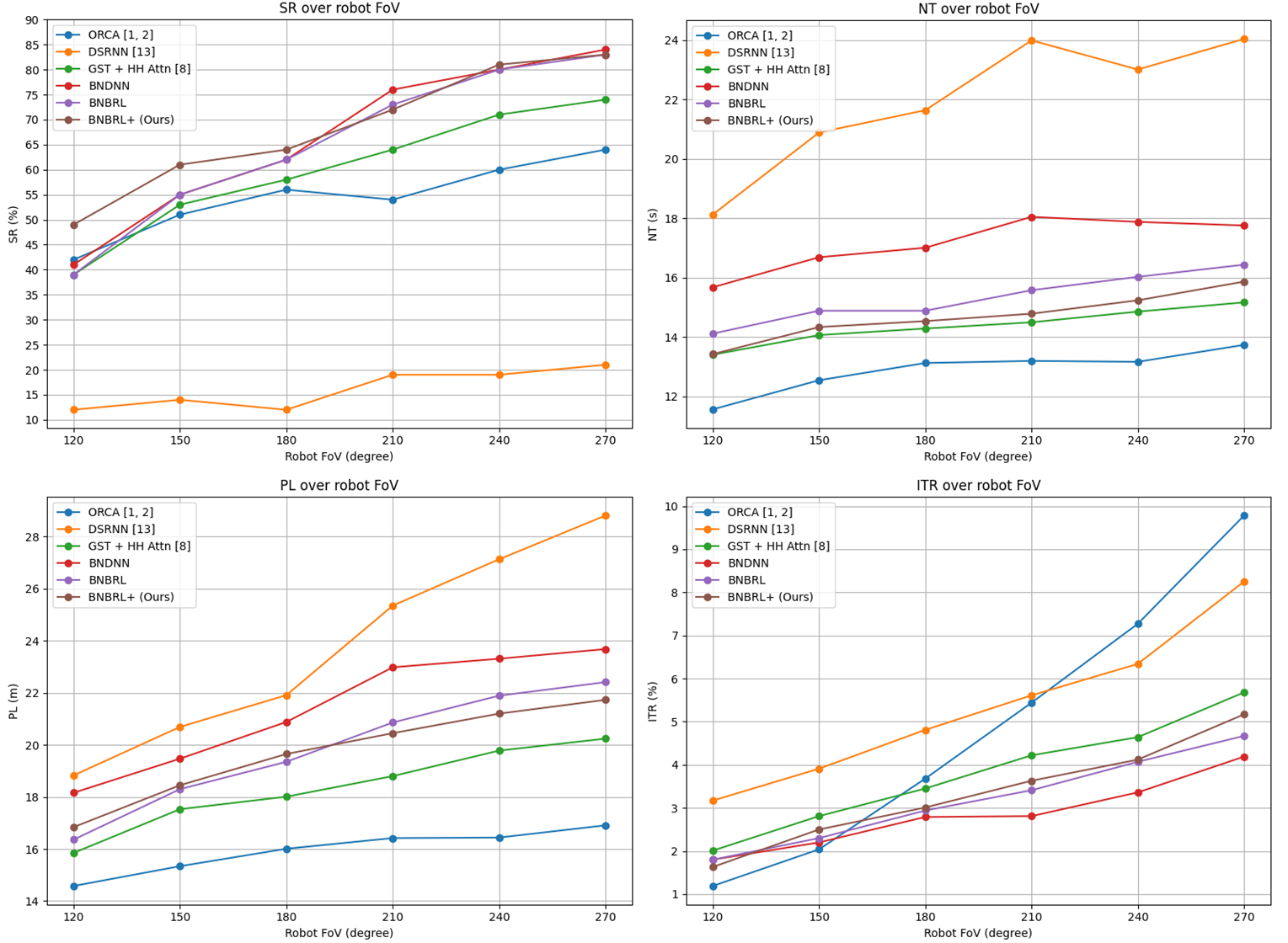}
      \caption{Evaluation metrics for decreasing FoV from 270° to 120° in increments of 30°}
      \label{figurelabel}
   \end{figure*}

As observed in Fig. 2, BNBRL+ secures a performance enhancement in the range 7–11\% over the three baseline models in terms of SR. Notably, GST+HH Attn is the highest-performing baseline at a FoV of 270° (73\%), which is surpassed by BNBRL+ with an 84\% performance, corresponding to an 11\% improvement. This result unequivocally establishes the superiority of BNBRL+ over other SOTA algorithms. Furthermore, BNBRL retains a performance on par with BNBRL+ from 270° to 180°, underscoring the significant role of belief and BNN in enhancing performance within POMDP environments.

Although the performance of BNDNN and BNBRL align with that of baseline models from 150° to 120°, BNBRL+ continues to exhibit an approximately 10\% advantage over competing algorithms. Therefore, the integration of belief and BNN via the attention mechanism strengthens its robustness and improves the performance even within the 150° to 120° range when compared to the DNN model. Consequently, BNBRL+ demonstrated high performance and robustness under constrained FoV conditions, highlighting the advantageous impact of belief algorithms, BNN, and attention modules.

As observed in Fig. 2, the metrics NT and PL provide insights on the robot's movement duration and traversed distance, respectively. These parameters typically exhibit a proportional relationship, reflecting similar trends. Thus, BNBRL+ is observed to either match or outperform DSRNN and GST+HH Attn, while significantly outperforming BNDNN and BNBRL. This aspect warrants analysis from both perspectives of efficiency and socially aware navigation. Increased NT and PL values indicate paths that are socially considerate (deemed safer by humans with minimal collision risk) but do not necessarily imply high efficiency. 

In contrast, reduced NT and PL values facilitate socially aware navigation but might lead to less efficient trajectories. Therefore, NT and PL are inversely related to efficiency and tend to correlate with the ITR metric, indicative of socially aware navigation. It is imperative to finely tune this balance to achieve optimal efficiency and SR. Based on this perspective, BNBRL+ consistently attains a high SR while ranking as the third most efficient model, exemplifying effective socially aware navigation alongside commendable performance and efficiency. 

\subsection{Result 2 (Blink)}

   \begin{figure*}[t]%
      \centering
      \includegraphics[width=1.0\textwidth]{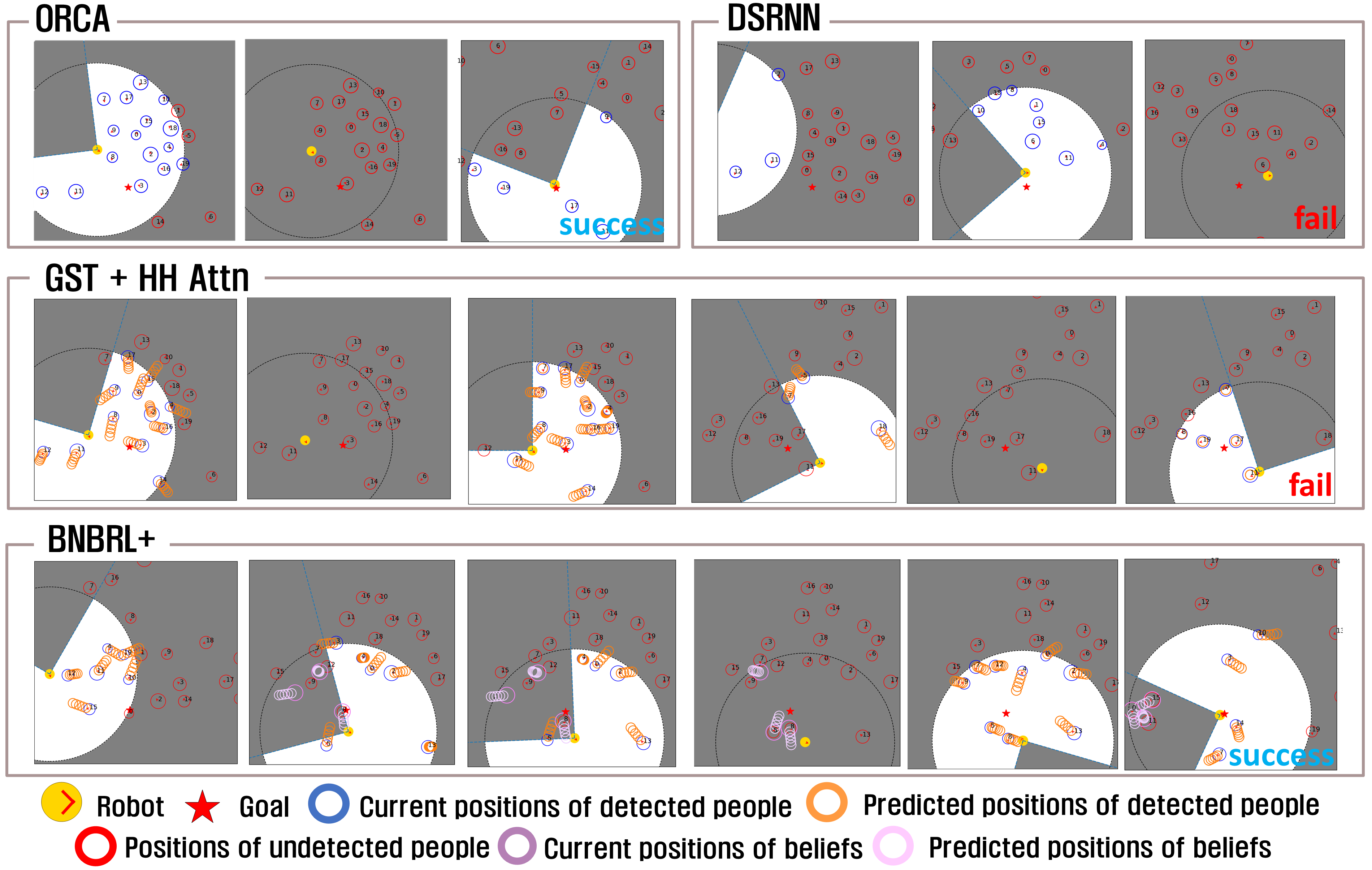}
      \caption{Comparison of driving paths for the same test scenario}
      \label{figurelabel}
   \end{figure*}

    \begin{table}[htb]
    \renewcommand{\arraystretch}{1.5}
    \caption{EXPERIMENTAL RESULTS IN BLINK SCENARIO}
    \resizebox{\columnwidth}{!}{%
    \begin{tabular}{cccccc}
    \hline
    Navigation Method  & SR($\Delta$) ($\%$) & NT($\Delta$) (s) & PL($\Delta$) (m) & ITR($\Delta$) ($\%$) \\
    \hline
    ORCA [1, 2]     & 63 (-1)  & \textbf{14.12 (+0.38)} & \textbf{16.8 (-0.11)}  & 8.24 (-1.54)  \\
    DRNN [13]       & 22 (+1)  & 24.16 (+0.12) & 26.26 (-2.55) & 6.61 (-1.64)  \\
    GST+HH Attn [8] & 63 (-11) & 14.37 (-0.8)  & 18.84 (-1.4)  & 5.49 (-0.19)  \\
    BNDNN           & 68 (-16) & 17.02 (-0.74) & 22.05 (-1.63) & \textbf{4.12 (-0.07)}  \\
    BNBRL           & 70 (-13) & 15.66 (-0.78) & 21.02 (-1.39) & 4.62 (-0.05)  \\
    BNBRL+ (Ours)   & \textbf{75 (-7)}  & 14.84 (-1.03) & 20.45 (-1.28) & 4.99 (-0.18)  \\
    \hline
    \end{tabular}
    }
    \end{table}

Robots may encounter scenarios where data are compromised or lost due to noise and physical constraints. Such instances are often misinterpreted by algorithms as either the non-presence of humans who are rapidly advancing toward their objectives or stopping due to errors. It is essential for robots to possess mechanisms that safeguard human safety in these situations.

The phenomenon of "blink" involves a scenario where a robot operates with a FoV of 270° for 3 time steps, followed by 0.5 time steps with an FoV of 0°, leading to intermittent disruptions in LiDAR data. This setup is utilized to assess the performance of the algorithm and robustness when the FoV of the robot alternates to 270°.

The trajectory and performance of the robot are depicted in Fig. 3. Although the ORCA method exhibited minimal performance fluctuations, it was marked by a significant ITR. As depicted in Fig. 3, the robot tends to approach humans during blink episodes. Nonetheless, owing to its rapid adaptation to changes in visibility, it manages to maintain a relatively high SR. However, its elevated ITR renders it less viable for socially aware navigation, with the reduction in ITR attributed to the index not escalating during blink episodes.

The DSRNN manifested a low SR alongside heightened NT and PL metrics. Moreover, an increase in $\Delta$NT while reduction in $\Delta$PL suggests the occurrence of the “freezing robot problem” owing to a lack of data resulting from failure to detect humans during blink situations. As portrayed in Fig. 3, DSRNN opts for considerably elongated routes to avert human collisions but frequently fails to reach the proximate destination, indicating ineffective learning due to inadequate human-related data. 

GST + HH Attn experienced an 11\% SR reduction, coupled with decreases in NT, PL, and ITR metrics. This outcome is likely influenced by the GRU, which facilitates safe and efficient navigation paths even in blink scenarios. 

Therefore, relatively low NT and PL indices are observed during blinking. Nonetheless, if an individual is positioned outside the FoV during a blink episode, the inability of the system to adequately react to the person leads to collisions, identified as the primary reason for the decline in SR. 

The BNDNN model exhibited a significant reduction of 16\%—a downturn attributed to the overconfidence dilemma intrinsic to DNN. Consequently, BNDNN excels in scenarios endowed with an expansive FoV, because of the negligible disparity between actual and perceived (belief) trajectories. Nonetheless, in instances where the FoV = 120–150° or belief inaccuracies increase, this overconfidence detrimentally affects the performance (Table 1). Therefore, integrating the BNN model significantly improved both performance and robustness.

The BNBRL+ ensemble recorded reductions in NT, PL, and ITR, alongside maintaining a relatively elevated SR with minimal SR degradation. As illustrated in Fig. 3, the robot adeptly interprets human movements to navigate safely and efficiently. Moreover, even when minimal numbers of individuals are positioned outside the FoV, similar to GST + HH Attn, or during blink episodes, BNBRL+ persistently maneuvers to evade the anticipated future paths of individuals based on belief. 

In contrast to other models, this distinct approach demonstrates its superior performance, ensuring successful arrival at the destination, which is not consistently realized by GST + HH Attn. Additionally, a comparative analysis with BNBRL highlights the effective synergy between the attention mechanism and BNN. 

\section{CONCLUSIONS}

This paper introduces an innovative methodology that integrates a belief algorithm within the POMDP framework and a BNN into DRL. The model showcased exceptional performance and robustness relative to other SOTA algorithms in challenging environments characterized by various risk factors, including limited FoV and intermittent visibility ("blink"). Consequently, this approach demonstrated considerable effectiveness, marked by certain constraints. However, our methodology relies heavily on trajectory prediction. Therefore, the algorithm is likely to be affected by the accuracy of the prediction algorithm. Therefore, future research is anticipated to reduce the dependence on the prediction algorithm and to delve into more nuanced scenarios.
   
\addtolength{\textheight}{-12cm}   % This command serves to balance the column lengths

%%%%%%%%%%%%%%%%%%%%%%%%%%%%%%%%%%%%%%%%%%%%%%%%%%%%%%%%%%%%%%%%%%%%%%%%%%%%%%%%

%%%%%%%%%%%%%%%%%%%%%%%%%%%%%%%%%%%%%%%%%%%%%%%%%%%%%%%%%%%%%%%%%%%%%%%%%%%%%%%%

%%%%%%%%%%%%%%%%%%%%%%%%%%%%%%%%%%%%%%%%%%%%%%%%%%%%%%%%%%%%%%%%%%%%%%%%%%%%%%%%

\section*{ACKNOWLEDGMENT}

This research was partly supported by the MSIT (Ministry of Science and ICT), Korea, under the Convergence security core talent training business support program (IITP-2023-RS-2023-00266615) supervised by the IITP (Institute for Information \& Communications Technology Planning \& Evaluation), and the BK21 plus program "AgeTech-Service Convergence Major" through the National Research Foundation (NRF) funded by the Ministry of Education of Korea [5120200313836], and Institute of Information \& communications Technology Planning \& Evaluation (IITP) grant funded by the Korea government (MSIT) (No.RS-2022-00155911, Artificial Intelligence Convergence Innovation Human Resources Development (Kyung Hee University)), and the Ministry of Trade, Industry and Energy (MOTIE), South Korea, under Industrial Technology Innovation Program under Grant 20015440, 20025094

%%%%%%%%%%%%%%%%%%%%%%%%%%%%%%%%%%%%%%%%%%%%%%%%%%%%%%%%%%%%%%%%%%%%%%%%%%%%%%%%


\begin{thebibliography}{99}

\bibitem{c1} Borenstein, J. and Y. Koren, Real-time obstacle avoidance for fast mobile robots. IEEE Transactions on systems, Man, and Cybernetics, 1989. 19(5): p. 1179-1187.
\bibitem{c2} Van Den Berg, J., et al. Reciprocal n-body collision avoidance. in Robotics Research: The 14th International Symposium ISRR. 2011. Springer.
\bibitem{c3} Helbing, D. and P. Molnar, Social force model for pedestrian dynamics. Physical review E, 1995. 51(5): p. 4282.
\bibitem{c4} Brown, N., Edward T. Hall: Proxemic Theory, 1966. Center for Spatially Integrated Social Science. University of California, Santa Barbara. http://www. csiss. org/classics/content/13 Read, 2001. 18: p. 2007.
\bibitem{c5} Mnih, V., et al., Playing atari with deep reinforcement learning. arXiv preprint arXiv:1312.5602, 2013.
\bibitem{c6} Wang, W., et al., Multi-robot cooperative socially-aware navigation using multi-agent reinforcement learning. arXiv preprint arXiv:2309.15234, 2023.
\bibitem{c7} Wang, R., W. Wang, and B.-C. Min. Feedback-efficient Active Preference Learning for Socially Aware Robot Navigation. in 2022 IEEE/RSJ International Conference on Intelligent Robots and Systems (IROS). 2022. IEEE.
\bibitem{c8} Liu, S., et al. Intention aware robot crowd navigation with attention-based interaction graph. in 2023 IEEE International Conference on Robotics and Automation (ICRA). 2023. IEEE.
\bibitem{c9} Bellman, R., The theory of dynamic programming. Bulletin of the American Mathematical Society, 1954. 60(6): p. 503-515.
\bibitem{c10} Mun, Y.-J., et al. Occlusion-aware crowd navigation using people as sensors. in 2023 IEEE International Conference on Robotics and Automation (ICRA). 2023. IEEE.
\bibitem{c11} Huang, Z., et al., Learning sparse interaction graphs of partially detected pedestrians for trajectory prediction. IEEE Robotics and Automation Letters, 2021. 7(2): p. 1198-1205.
\bibitem{c12} Omidshafiei, S., et al. Decentralized control of partially observable markov decision processes using belief space macro-actions. in 2015 IEEE international conference on robotics and automation (ICRA). 2015. IEEE.
\bibitem{c13} Liu, S., et al. Decentralized structural-rnn for robot crowd navigation with deep reinforcement learning. in 2021 IEEE International Conference on Robotics and Automation (ICRA). 2021. IEEE.
\bibitem{c14} Blundell, C., et al. Weight uncertainty in neural network. in International conference on machine learning. 2015. PMLR.
\bibitem{c15} Chen, C., et al. Relational graph learning for crowd navigation. in 2020 IEEE/RSJ International Conference on Intelligent Robots and Systems (IROS). 2020. IEEE.
\bibitem{c16} Trautman, P. and A. Krause. Unfreezing the robot: Navigation in dense, interacting crowds. in 2010 IEEE/RSJ International Conference on Intelligent Robots and Systems. 2010. IEEE.
\bibitem{c17} Fu, R., Z. Zhang, and L. Li. Using LSTM and GRU neural network methods for traffic flow prediction. in 2016 31st Youth academic annual conference of Chinese association of automation (YAC). 2016. IEEE.
\bibitem{c18} Bouton, M., et al. Safe reinforcement learning with scene decomposition for navigating complex urban environments. in 2019 IEEE Intelligent Vehicles Symposium (IV). 2019. IEEE.
\bibitem{c19} Sun, L., et al. Behavior planning of autonomous cars with social perception. in 2019 IEEE Intelligent Vehicles Symposium (IV). 2019. IEEE.
\bibitem{c20} Nguyen, H., et al. On-Robot Bayesian Reinforcement Learning for POMDPs. in 2023 IEEE/RSJ International Conference on Intelligent Robots and Systems (IROS). 2023. IEEE.
\bibitem{c21} Png, S.C.O.S.W. and D.H.W.S. Lee. PODMPs for robotic tasks with mixed observability. in the Conference Proceedings of Robotics: Science and Systems, Seattle, WA. 2009.
\bibitem{c22} Littman, M.L., A.R. Cassandra, and L.P. Kaelbling, Learning policies for partially observable environments: Scaling up, in Machine Learning Proceedings 1995. 1995, Elsevier. p. 362-370.
\bibitem{c23} Konda, V. and J. Tsitsiklis, Actor-critic algorithms. Advances in neural information processing systems, 1999. 12.
\bibitem{c24} Niu, Z., G. Zhong, and H. Yu, A review on the attention mechanism of deep learning. Neurocomputing, 2021. 452: p. 48-62.
\bibitem{c25} Schulman, J., et al., Proximal policy optimization algorithms. arXiv preprint arXiv:1707.06347, 2017.
\bibitem{c26} Kim, J., et al., Transformable Gaussian Reward Function for Socially-Aware Navigation with Deep Reinforcement Learning. arXiv preprint arXiv:2402.14569, 2024.
\bibitem{c27} Jia, D., A. Hermans, and B. Leibe. DR-SPAAM: A spatial-attention and auto-regressive model for person detection in 2D range data. in 2020 IEEE/RSJ International Conference on Intelligent Robots and Systems (IROS). 2020. IEEE.

\end{thebibliography}
\end{document}